\documentclass[conference]{IEEEtran}
\IEEEoverridecommandlockouts
\usepackage{cite}
\usepackage{amsmath,amssymb,amsfonts}
\usepackage{algorithmic}
\usepackage{graphicx}
\usepackage{textcomp}
\usepackage{xcolor}
\usepackage{array}
\def\BibTeX{{\rm B\kern-.05em{\sc i\kern-.025em b}\kern-.08em
    T\kern-.1667em\lower.7ex\hbox{E}\kern-.125emX}}
\begin{document}

\title{Application of Autoencoder-Assisted Recurrent Neural Networks to Prevent Cases of Sudden Infant Death Syndrome}
\author{\IEEEauthorblockN{Maximilian Du}
\IEEEauthorblockA{\textit{Fayetteville-Manlius High School} \\
Manlius, NY, United States\\
maxressb@gmail.com}}
\maketitle
\begin{abstract}
This project develops and trains a Recurrent Neural Network (RNN) that monitors sleeping infants from an auxiliary microphone for cases of Sudden Infant Death Syndrome (SIDS), manifested in sudden or gradual respiratory arrest. To minimize invasiveness and maximize economic viability, an electret microphone, and parabolic concentrator, paired with a specially designed and tuned amplifier circuit, was used as a very sensitive audio monitoring device, which fed data to the RNN model. This RNN was trained and operated in the frequency domain, where the respiratory activity is most unique from noise. In both training and operation, a Fourier transform and an autoencoder compression were applied to the raw audio, and this transformed audio data was fed into the model in 1/8 second time steps. In operation, this model flagged each perceived breath, and the time between breaths was analyzed through a statistical T-test for slope, which detected dangerous trends. The entire model achieved 92.5\% accuracy on continuous data and had an 11.25-second response rate on data that emulated total respiratory arrest. Because of the compatibility of the trained model with many off-the-shelf devices like Android phones and Raspberry Pi's, free-standing processing hardware deployment is a very feasible future goal. 
\end{abstract}
\maketitle
\section{Problem Formulation}
Sudden Infant Death Syndrome (SIDS) affected 1500 people in the United States in 2016 \cite{SIDS}. This fatal condition has not been narrowed down to a single cause, but there have been linkages to brain defects within portions of the brain that controls breathing and sleep \cite{Brain}. As such, although SIDS is multifarious in the immediate cause, one common manifestation of this condition is an irregular respiratory activity that eventually leads to total respiratory failure. 

Although monitoring devices exist for SIDS and other infant conditions, the majority are either expensive, invasive, or slow in response. Some examples of existing devices include, a blood oximeter, which is slow to react since blood oxygen levels can remain tolerable after respiratory failure \cite{Oxygen},  an accelerometer-based clip system, which slightly invasive \cite{Clip}, and a chest impedance-based respiratory monitor, which is under development and expensive, but remains the least invasive and fastest reacting option \cite{Impedance}. 

Although auditory monitoring of infants has been widely used, most simply act as intercoms without processing the sounds. By using artificial intelligence to process the sounds of the infant, a more autonomous and minimally invasive monitor can be made, since the artificial intelligence will flag each breath and also monitor the frequency of breaths for irregularities, all without constant human surveillance.
\section{Background}
Artificial intelligence takes many forms, but artificial neural networks were decided to be the best approach to autonomous monitoring. Although simple feed-forward neural networks are powerful in classification or prediction, it does not deal with temporal data very accurately, and audio is temporal in many respects. 

Thus, a Recurrent Neural Network (RNN) was best fit for sound detection. An RNN recursively propagates hidden layers forward so that its current state contains some representation of all past inputs \cite{RNN}. Since inputs farther in the past have hidden layers convoluted sequential operations, the network essentially has a directional memory. The more recent inputs have a higher impact on the output, which mirrors how natural memory works. Advanced forms of RNNs include LSTMs (Long Short-Term Memory) and GRUs (Gated Recurrent Unit), which use sub-neural networks to control memory decay \cite{RNN}. 

\section{Hardware}
\begin{figure}[htb]
\centering
\includegraphics[width=0.50\textwidth]{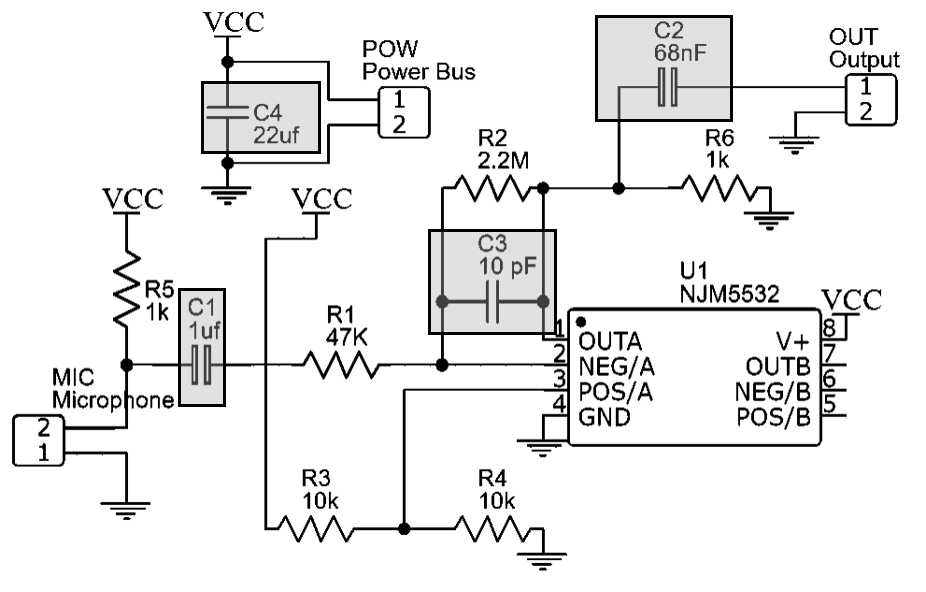}
\caption{Amplifier Circuit (Boxes are noise reduction filter components)}
\label{fig:circuit}
\end{figure}

\begin{figure}[htb]
\centering
\includegraphics[width=0.25\textwidth]{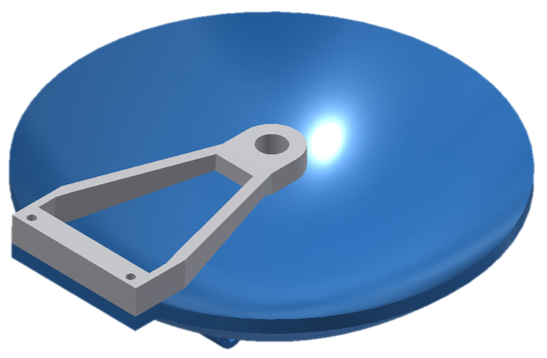}
\caption{Elliptical Paraboloid Concentrator and Microphone Arm}
\label{fig:parabola}
\end{figure}

While monitoring microphones are widely available in the consumer market, a custom assembly was made because economic consumer options are not sensitive enough to accurately record respiratory activity, and professional options would reduce availability to potential consumers. The design consisted of an electret microphone at the focal point of a 15 cm diameter paraboloid (Fig~\ref{fig:parabola}) which acted as a passive amplifier by concentrating a large area of sound to a single point. While this assembly was functional, it lacked a suitable driver circuit that would both amplify and perform noise reduction. In order to address both issues, a custom amplifier circuit and high/low pass filters were fabricated and tuned (Fig.~\ref{fig:circuit}). The filters targeted both ambient noise from the microphone and induced EMF noise. The output of this device was in the form of an analog audio signal which was processed into a digital signal by a sound card, which was fed into the model.

\section{Data Collection and Parsing}
Due to the inability to procure true infant breathing data, substitute breathing data, both from open source data sets and recorded from non-infants were used. A large number (1500 clips) of high variety samples prevents overfitting.

Although a higher recording resolution could have been used, such higher data rates increase computational complexity for the model and also increase the risk of overfitting the model. After experiments with varying sampling rates, an 8192 bit/s sampling rate was the lowest sampling rate in which noise did not take up significant bandwidth. According to the Nyquist limit, the maximum measurable frequency with this rate was 4096-hertz \cite{Nyquist}. This frequency cutoff meant that roughly 75\% of frequencies present respiratory sounds were accounted for with this sample rate. 

Each audio clip was two seconds long, which spans a typical breath, and was part of a balanced set with three classes: inhale, exhale, and unknown. Each class had 500 samples, and the balancing prevented prediction bias in the final model. Although inhale and exhale could have been concatenated, the model was made with future developments in mind where inhale and exhale patterns will be treated as distinct.  

\begin{figure}[htb]
\centering
\includegraphics[width=0.5\textwidth]{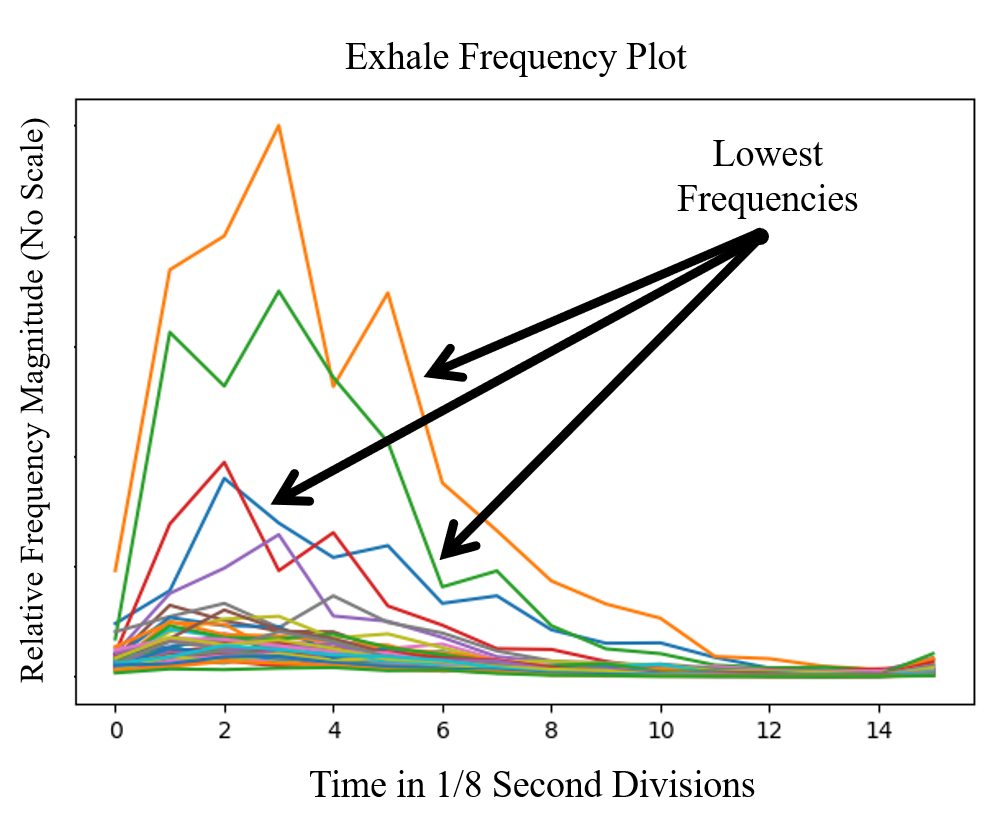}
\caption{Frequency Plot for an Exhale. Note the prominent lower frequencies}
\label{fig:plot}
\end{figure}

Although simple amplitude points could have been used as the inputs to the RNN, the respiratory activity that was examined had no distinct pattern from ambient noise in the amplitude domain, but the same respiratory activity had pronounced frequencies that made its plot in frequency domain distinct (Fig~\ref{fig:plot}). As such, the input audio was pre-processed using a discrete fast Fourier transform (DFFT), a variation of the Fourier transform function. To accomplish this, an algorithm splits each two-second sample into sections 1/8 second long. Each of these sections was independently transformed into the frequency domain, giving a frequency plot over time.

In a separate training program, three hundred training samples were chosen at random from the pool of 1400 per epoch. The 100 excluded clips were used in testing and validation. 50 clips were chosen at the beginning of training and isolated for testing after training. The other half was dynamically randomly chosen from the large pool of data. This half served as the validation set to give an evaluation on how the network performs during training. 

\section{Frequency Compression with Autoencoder}
The DFFT yielded a 1024-value vector per 1/8 second interval. While this could have been the input to the RNN, this amount of data would result in higher training computational complexity and high levels of overfitting. There were many methods to reduce the dimensions of the DFFT, including principal component analysis (PCA) and t-SNE reduction, but the optimal structure was found to be an autoencoder due to its higher adaptability. This structure exploits a deep perceptron network structure, learning a lesser representation of the input data by training it to reconstruct the input data from a lower dimension hidden layer(Fig 4). As empirically derived, a compression from 1024 to 50 dimensions was possible without significant losses, and an optimized autoencoder capable of this compression was used between the DFFT output and the RNN input. The utilization of an autoencoder resulted in feasible training times and increased the robustness of the network in handling frequency shifts. 
\begin{figure}[htb]
\centering
\includegraphics[width=0.5\textwidth]{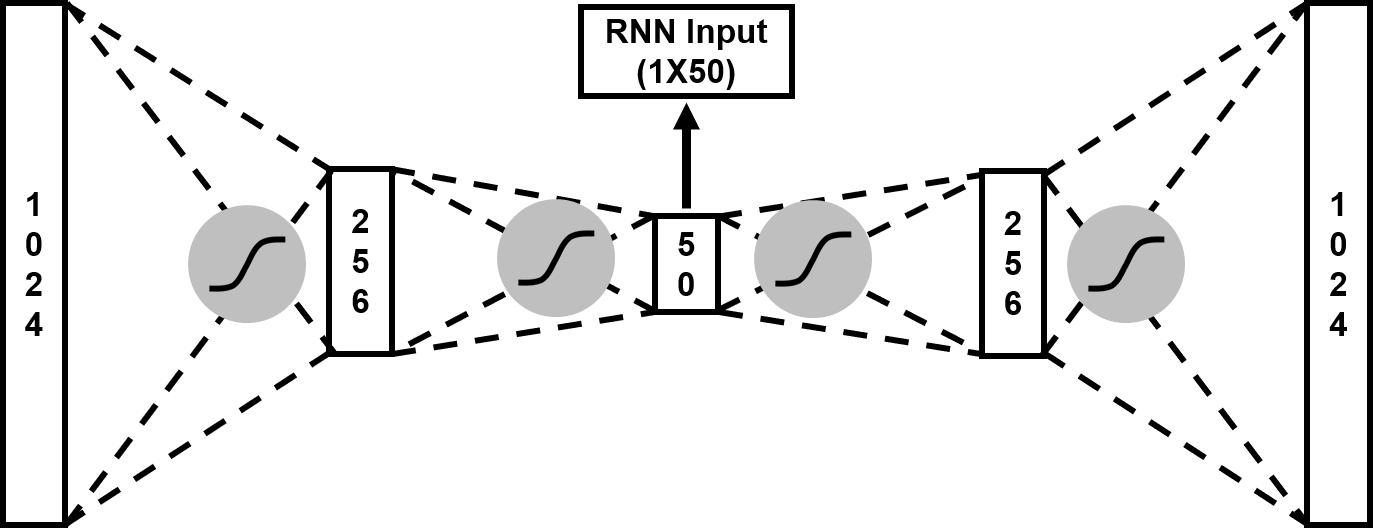}
\caption{Autoencoder Structure}
\label{fig:autoencoder}
\end{figure}
\section{Model Structure of RNN and Training}
Although the RNN has many variants like the LSTM and GRU, a simple, non-gated RNN was chosen because of the simplicity of the data. In the frequency domain, an inhales, exhales, and background noises were very distinct and there was no need for more degrees of training freedom. 
\subsection{RNN Structure}
\begin{figure}[htb]
\centering
\includegraphics[width=0.5\textwidth]{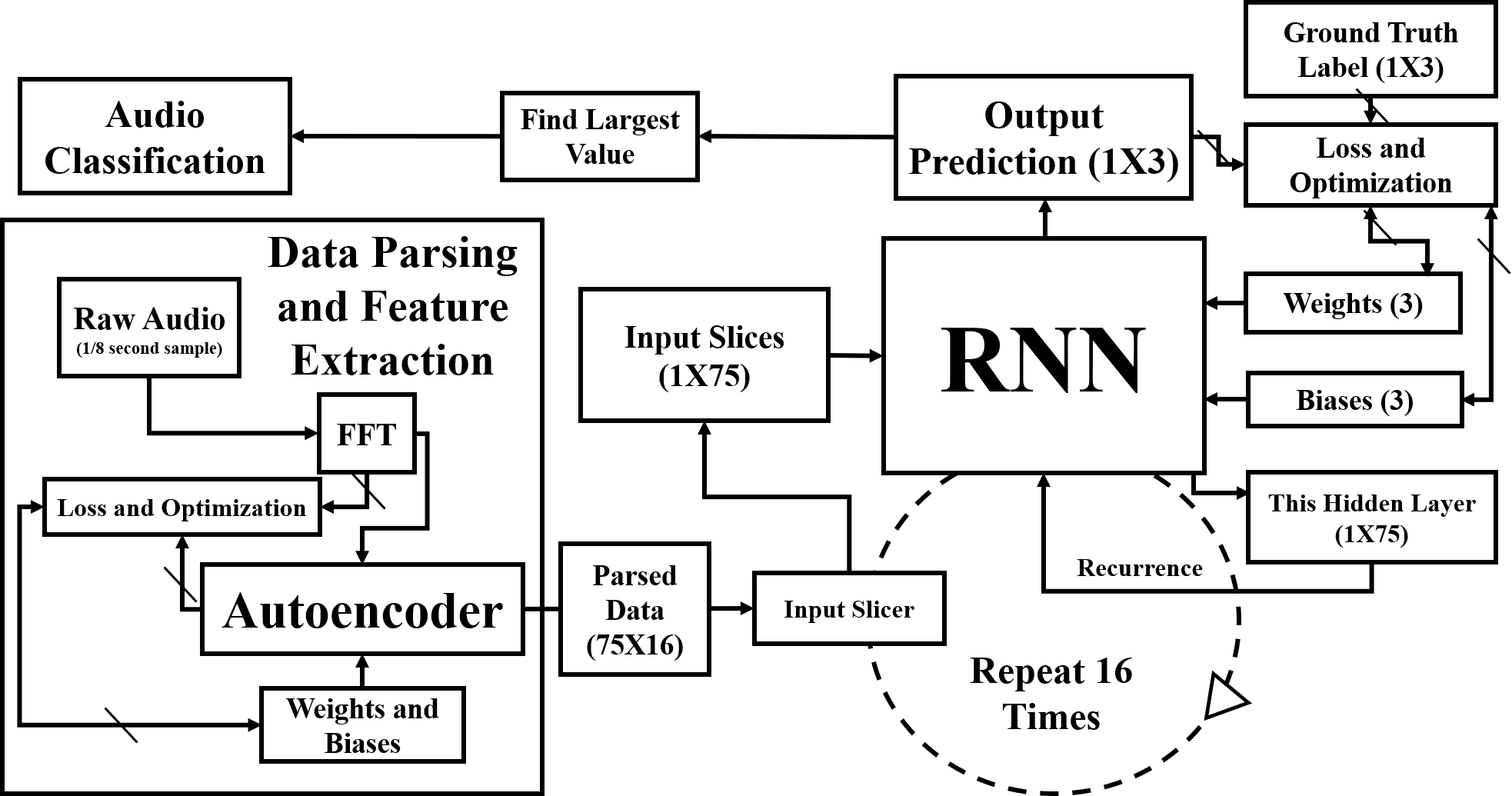}
\caption{RNN Data Diagram. Arrows represent data flow. Dotted lines represent control sequences. Slashed lines represent control sequences. Slashed lines represent training-only operations}
\label{fig:RNN}
\end{figure}

Within the basic RNN model, there were many factors that were considered, and often the optimal options were empirically derived. The final RNN model structure was a single, hyperbolic tangent activation function-based hidden layer structure with a many-to-one architecture (Fig~\ref{fig:RNN}) and was trained with the Adagrad optimizer. It is important to note that the hyperbolic tangent function was used to prevent exploding gradient and to allow for negative values within the model, but the output layer had a sigmoid activation function in place of the hyperbolic tangent because prediction confidences are always positive.
\subsection{Network Optimization}
While the RNN structure was relatively simple to optimize, an optimized network took more computational power due to more degrees of freedom. In the search for an optimized network, five models were trained, each with different hyperparameters. In the first three versions, overfitting was observed, but not from standard tests for overfitting involving the untrained test set. Confusion matrixes and F1 scores were incredibly promising, but the trained model became chaotic when tested with sounds recorded in slightly different ambiances. 

While the network remained optimal for the audio data set as a whole, a large number of hidden neurons (100) allowed the RNN to capture small details, including small variations in noise. A larger number of hidden neurons allowed for more processing depth but also allowed for false insights on trivial aspects of data. Simple variations in background noise would make the model predict "unknown" with more than 99\% confidence in breathing clips. 

Thus, for the fourth and fifth versions, the hidden neuron size was modified. The fourth model used only fifty hidden neurons, which unfortunately resulted in slow convergence and low accuracy. However, for the firth model, new audio samples, augmented by the addition of random noise, were fed into a model with 75 hidden neurons. This number of neurons, coupled with noise augmentation, made an optimal model. It scored an F1 score of 0.97 and had no false predictions on a confusion matrix. More importantly, it also remained robust in varied conditions, even remaining above 81\% accurate under different microphones.
\subsection{Coding and Development}
The entire model was programmed in Python with the Tensorflow library by Google. Although pre-made RNN models were available in the library, none were used, so that structural adjustment and debugging were made easier. To train the model, an Nvidia GTX 1060 was used to train 20,000 epochs per model.  

\section{Statistical Analysis of Predictions on Continuous Audio Signals}
\subsection{Continuous Operation}
This RNN was trained on discreet, 2-second samples, but in operation, it must run continuously, but more importantly, be able to catch all breaths and report no false positives. Many approaches were considered, but the best option was through overlapping model instances. By running a new instance of the model every 1/8 second on the continuous data, there will always be at least one window in which the inhale or exhale was completely captured because each shift represents the smallest resolution of the RNN. 
While this model was very successful in recognition of audio, it also produced many false positives in the time interval leading up and after the audio event. A filtering algorithm, which ignored all predictions below 99\% confidence and required three sequential positives, was able to provide a clean timestamp for every breath. 
\subsection{Irregularity Testing}
In order to test for irregularities, two factors were considered: irregular respiratory activity and total respiratory arrest. Total respiratory arrest has a straightforward implementation: an algorithm finds the time between breaths and sets up an 80\% confidence interval for acceptable times. If the current time exceeds the interval, the alert is triggered.

To detect irregular respiratory activity, the same data, the time between breaths, was used. Instead of a confidence interval, a T-test for a non-zero slope was used. This allows prolonged and dangerous trends to be detected. 

The reason why statistical tests were used lies in individual variation. No single threshold can be set for all infants, but by using the infant's own past data, any outliers are true outlets to the infant's own behavior, eliminating the problems that a set threshold gives, which include false alarms and ignored true conditions. In addition, a statistical test is less sensitive to random sensor and RNN model noise, since the T-test and confidence interval look holistically. 

\section{Results}
\begin{figure}[htb]
\centering
\includegraphics[width=0.35\textwidth]{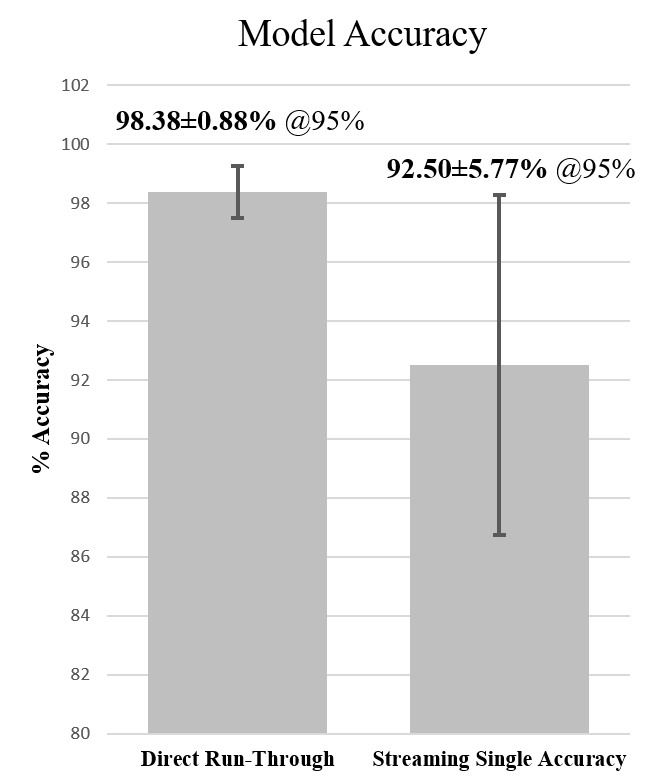}
\caption{RNN Accuracy}
\label{fig:RNNacc}
\end{figure}

There were many measures of accuracy on this entire model, ranging from RNN accuracy testing to whole-model condition emulation. 
In direct, discreet RNN performance, the trained network had an F1 score of 0.97 and could classify $98.38 \pm 0.88\%$ of sounds accurately (Figure \ref{fig:RNNacc})

In continuous RNN model performance, the model was able to accurately distinguish $92.5 \pm 5.77\%$ of breaths from background noise within one second of playback. This test was done on a constructed breath sequence using untrained data and genuine ambient noise. More importantly, it had no false positive breath detections, which is consistent with the purpose of the model, as a false positive could overlook respiratory arrest.

\begin{figure}[htb]
\centering
\includegraphics[width=0.5\textwidth]{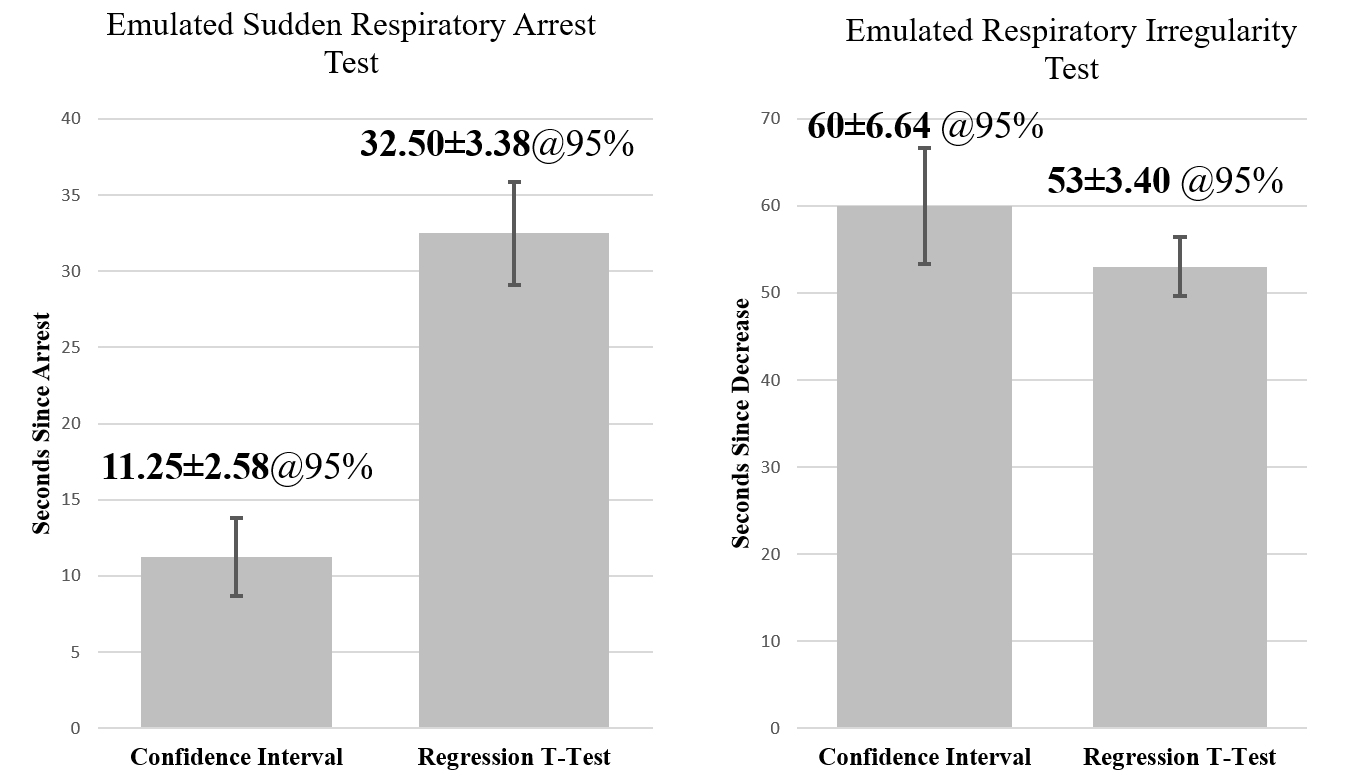}
\caption{Emulated Whole-System Test Results}
\label{fig:RespArr}
\end{figure}

In a whole-model condition emulation, the combined T-test and confidence interval algorithm was able to detect complete respiratory arrest within $11.25 \pm 2.58$ seconds and respiratory rate decrementation
within $53.00 \pm 3.40$ seconds of its commencement (Figure~\ref{fig:RespArr}). 

It is seen that both confidence interval and T-test detection algorithms are integral, just in different situations. The confidence interval was able to detect a direct respiratory arrest far quicker than the T-test could, but the T-test was able to detect the emulated respiratory irregularities quicker than the confidence interval could, proving that both modes of detection must work in tandem. 

\section{Conclusions and Future Work}
While infant breathing data was not used in the project, it is reasonable to say that this method of respiratory monitoring can be used in such a way. With a reaction time comparable to accelerometer-based clip-on systems and minimal additional hardware, this monitoring system may be a feasible alternative to current market models, but a large amount of follow-up work is needed to support this claim. 

Future development is very multifaceted, but the most notable development is the utilization of standalone processors connected to the internet like the Raspberry Pi to run the trained model and statistical algorithms, allowing monitoring from anywhere around the world. In addition, real infant breathing data would allow for a true accuracy assessment of the model. 

\bibliographystyle{unsrt}
\bibliography{ref}

\end{document}